# PFML-based Semantic BCI Agent for Game of Go Learning and Prediction

Chang-Shing Lee, Mei-Hui Wang
Dept. of Computer Science and Information Engineering
National University of Tainan
Tainan, Taiwan
leecs@mail.nutn.edu.tw
mh.alice.wang@gmail.com

Li-Wei Ko, Bo-Yu Tsai
Institute of Bioinformatics and Systems Biology, Center For Intelligent Drug Systems and Smart Bio-devices (IDS2B)
National Chiao Tung University
Hsinchu, Taiwan
lwko@mail.nctu.edu.tw
p840507@gmail.com

Yi-Lin Tsai, Sheng-Chi Yang
Dept. of Computer Science and Information Engineering
National University of Tainan
Tainan, Taiwan
andy6804tw@gmail.com
skymini4910@gmail.com

Lu-An Lin
Taiwan Go Association
Taipei, Taiwan
luan20050427@gmail.com

Yi-Hsiu Lee, Hirofumi Ohashi
Japan Go Association
Tokyo, Japan
good-luck0616@hotmail.co.jp
xsp7top@yahoo.co.jp

Naoyuki Kubota, Nan Shuo
Dept. of System Design
Tokyo Metropolitan University
Tokyo, Japan
kubota@tmu.ac.jp
shuo-nan@ed.tmu.ac.jp

*Abstract*—This paper presents a semantic brain computer interface (BCI) agent with particle swarm optimization (PSO) based on a Fuzzy Markup Language (FML) for Go learning and prediction applications. Additionally, we also establish an Open Go Darkforest (OGD) cloud platform with Facebook AI research (FAIR) open source Darkforest and ELF OpenGo AI bots. The Japanese robot Palro will simultaneously predict the move advantage in the board game Go to the Go players for reference or learning. The proposed semantic BCI agent operates efficiently by the human-based BCI data from their brain waves and machine-based game data from the prediction of the OGD cloud platform for optimizing the parameters between humans and machines. Experimental results show that the proposed human and smart machine co-learning mechanism performs favorably. We hope to provide students with a better online learning environment, combining different kinds of handheld devices, robots, or computer equipment, to achieve a desired and intellectual learning goal in the future.

*Keywords—particle swarm optimization, fuzzy markup language, brain computer interface, game of Go, FAIR ELF OpenGo*

## I. INTRODUCTION

Inspired by DeepMind's work (AlphaGoZero) [15], Facebook AI research (FAIR) reproduced and released the ELF OpenGo AI bot that is able to teach itself how to play Go at the level of a professional human player or better [13]. In this paper, we present a brain computer interface (BCI) agent with particle swarm optimization (PSO) based on a fuzzy markup language (FML) for Go learning and prediction applications. An Open Go Darkforest (OGD) cloud platform, including Facebook AI Research (FAIR) open sourced Darkforest [11] and ELF Open Go AI bots [13], is established to predict next top five moves for both Black and White; and the robot Palro, developed by Fujisoft, Japan, will predict current game situation and the next move advantage for Black or White as a reference during the competition [11, 12, 14]. Different humans may take different kinds of strategies even for an identical situation with their mood swinging. They may give a stable or an aggressive response to their opponent in a complicated game situation. Consequently, we want to further observe the variance of brain waves of the human who is playing and infer the linguistics of his/her each-move win rate based on the information extracted from humans and machines.

Brain is a very complex part of human body and it is also the center of all thoughts and life [1]. Brain signaling has emerged as a powerful candidate of the existing biometric traits due to its unique nature [17]. In the neuroscience field, there has been an increasing interest in studies about mapping the human brain connectivity in recent years [3]. Brain computer interface (BCI) is a bridge between the brain waves and the machine that can put the produced signals into effect [2]. There have been a wide range of successful applications about BCI [3]. For example, Martinez-Cagigal et al. [5] presented an asynchronous P300-based BCI system for controlling social networking features of a smartphone. Sitaram et al. [16] built an online support vector machine (SVM) to assess emotional disorders from fMRI signals. Lin et al. [4] estimated shifts in drivers' levels of arousal, fatigue, and vigilance based on a developed wireless and wearable electroencephalographic (EEG) system. Ko et al. [26] investigated students' sustained attention from alertness to fatigue in the real classroom via EEG activities changes. Above studies showed that many BCI systems have been widely developed for various applications and close to our real life.

MarketsandMarkets [6] reported that learning environments with the use of the AI technology simulate students passing for learning and help enhance their learning experience. AI can present information and provide practice time, without

The authors would like to thank the financially support sponsored by the Ministry of Science and Technology of Taiwan under the grants MOST 107-2218-E-024-001, and in part supported by the "Center For Intelligent Drug Systems and Smart Bio-devices (IDS2B)" from The Featured Areas Research Center Program within the framework of the Higher Education Sprout Project by the Ministry of Education (MOE) in Taiwan.

becoming impatient or judgemental [7]. In recent years, advances in technology have already transformed robots into the ones to co-learn with humans [7, 9, 10]. Lee et al. [10] proposed a machine-human co-learning model to help various students learn the mathematical concepts based on their learning ability and performance. Meanwhile, the robot acts as a teacher's assistant to co-learn with children in class. Lee et al. [8] also proposed a novel PSO-based FML (PFML) learning mechanism for optimizing the parameters between items and students based on item response theory (IRT) and a human fuzzy linguistic knowledge cognition model for future educational applications.

Fuzzy markup language (FML), an IEEE 1855-2016 standard, facilitates the modelling of a fuzzy controller in a human-readable and hardware-independent manner [18-19]. Considerable research has focused on FML applications, including computer games [21], diet [22], and student performance learning [8]. Training data, knowledge, and learning process are the important parts of AI technology. With the learned model and explainable model with argumentation, we can generate explanations for the output [20]. This paper proposes a PFML-based semantic BCI agent and the application of Go learning with prediction. We use the extracted features from human's brain waves and predicted features from machines as the training data. Additionally, we adopt FML to describe the human-readable knowledge. After that, we adopt the learning process based on PSO-based FML optimization to generate the learned model to infer human each-move linguistics. The remainder of this paper is organized as follows: Sections II and III introduce the implemented BCI agent and PFML learning for semantic BCI agent, respectively. The experimental results are shown in Section IV and conclusions are given in Section V.

## II. BRAIN-COMPUTER-INTERFACE (BCI) AGENT

### A. Introduction to Human Brain Waves

EEG is an electrophysiological monitoring method to record electrical activity of human brain [23]. In this study, we adopted the commercial EEG headset with 8 channels to collect the Go players' EEG signals when they are playing. Fig. 1 (left) shows the eight channels' location followed by the international 10-20 systems and Fig. 1 (right) is the adopted commercial EEG system photo called BR8 [24]. Each EEG channel with letters "F, T, P, and O" identifies the frontal, temporal, parietal, and occipital lobes, respectively.

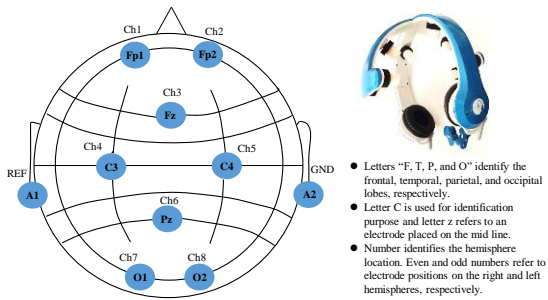

Fig. 1. Location of the eight channels and picture of the adopted mindo [24].

After collecting the EEG signals, we perform the time-frequency analysis to transfer the time domain EEG signals to different frequency brain waveforms. Normally, brain waveforms can be subdivided into bandwidths known as gamma, beta, alpha, theta, and delta whose descriptions are given as follows: 1) Gamma waves (30–80 Hz) have been linked to states of high attention. 2) Beta waves (12–30 Hz) may be involved in movement and complex tasks such as memory and decision making. 3) Alpha waves (8–12 Hz) appear when a relaxed person closes his eyes. 4) Theta waves (4–8 Hz) may help the brain sort information essential for navigation. 5) Delta waves (1.5–4 Hz) mark deep sleep and anesthesia [23].

### B. Brain-Computer-Interface Agent Structure

Fig. 2 shows the BCI agent structure. The agent is used to retrieve the human-based BCI data and machine-based game data during playing the game of Go with FAIR ELF OpenGo and Darkforest AI bot. We briefly describe the operation of the proposed structure as follows:

1) A human wearing a wireless BR8 [4, 24] plays Go with machines via the OGD cloud platform. The BCI-based psychological signal detection mechanism and the Go robot agent communicate with the OGD cloud platform via websocket.

2) When the human plays a move, the Go robot agent reports him/her a predicted move advantage and the current game situation with short linguistic description, including: *Black/White may be at a disadvantage*, *The winner still hasn't been determined*, *Black/White is at an advantage*, or *Black/White may win*, when the current game situation has been changed.

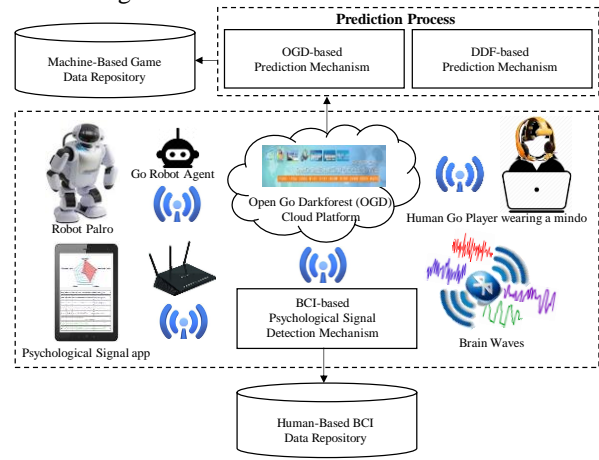

Fig. 2. BCI agent structure.

3) The BCI-based psychological signal detection mechanism continuously receives his/her brain waves via Bluetooth and then analyzes periodic rhythms recorded in the EEG. After that, we can acquire five types of indicators including: the human's attention level, left-brain activation level, right-brain activation level, stress level, and fatigue level. The developed psychological signal app shows and updates these five indicators about each second. Simultaneously, the EEG signal, the analyzed five indicators, and playing-time of each move are also stored in the human-based BCI data repository.

4) During playing, the OGD-based prediction mechanism and the DDF-based prediction mechanism predict the next top five moves based on ELF OpenGo [13] and Darkforest [11] AI engines, respectively. Each predicted move includes its position, simulation numbers, and win rate. These predicted data are stored into the machine-based game data repository.

### C. BCI-based Psychological Signal Detection Mechanism

The BCI-based psychological signal detection mechanism is responsible for analyzing the brain waves from 8 channels of the mindo and transferring them into five indicators, including attention, left-brain activation level, right-brain activation level, stress, and fatigue. Fig. 3 shows the information which we analyze the indication of the brain activity [25].

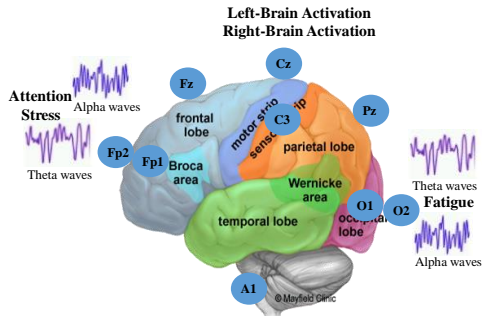

Fig. 3. Related informatio of the brain and channel locations [25].

1) Attention level: The alpha-band energy of the frontal lobe has a significant decreasing trend when a person is in high concentration, which is in contrast to the rest state. In this paper, we analyze the alpha waves of the channels Fp1, Fp2, and Fz to evaluate the attention level of the Go player.
2) Left-brain activation level and right-brain activation level: When the left brain is activated, the right brain will be negatively correlated with the left brain. We use the brain waves of the channels C3, C4, and Pz to measure these two corresponding indicators.
3) Stress level: We evaluate a person's stress indicator by the changes in theta and alpha bands of frontal lobe, including channels Fp1, Fp2, and Fz.
4) Fatigue level: The fatigue is highly associated with the energy of the occipital alpha and theta bands of channels O1 and O2 based on the past EEG studies [4, 26]. When the degree of fatigue level is increasing, the corresponding EEG power in occipital area is increasing significantly.

### III. PFML LEARNING FOR SEMANTIC BCI AGENT

#### A. PFML optimization Structure for Semantic BCI Agent

Fig. 4 shows the structure used to integrate the OGD cloud platform with PFML optimization for Go learning and prediction. We briefly describe the operation of the proposed structure as follows:

1) Go players play Go via the OGD cloud platform to generate the human-based BCI data. At the same time, the OGD cloud platform also predicts the information of the next five moves to create machine-based game data.
2) According to the machine-based game data and human-based BCI data, the domain expert constructs the knowledge base (KB) and rule base (RB) of the semantic BCI agent and stores the personalized BCI with knowledge about Go into the KB/RB repository.

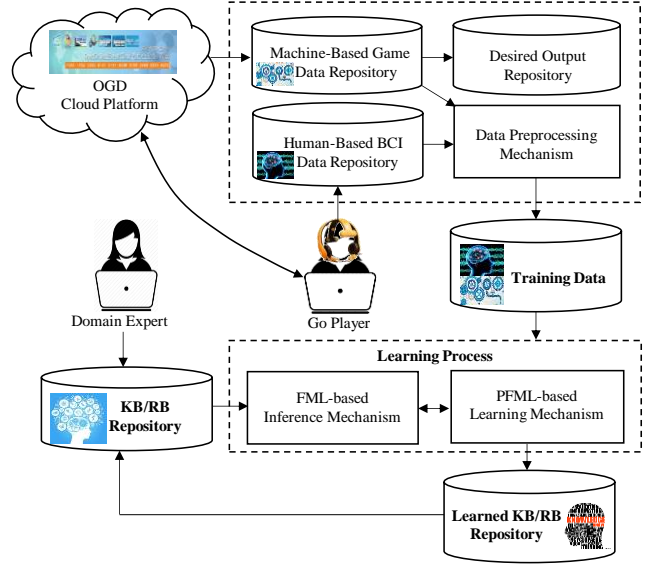

Fig. 4. PFML learning structure for semantic BCI agent.

3) The data preprocessing mechanism analyzes the human-based BCI data and machine-based game data to generate the training data. We also adopt the FAIR ELF OpenGo-predicted win rate as the desired output (DO) of the semantic BCI agent.
4) The PFML learning process, including an FML-based inference mechanism and a PFML-based learning mechanism, employs reasoning based on the learned KB and RB until termination and stores the parameters of the learned model into the learned KB/RB repository. After that, the learned KB/RB is feedback to the KB/RB repository, and this can facilitate human-and-machine co-learning.

#### B. Data Preprocessing Mechanism

This section introduces the data preprocessing mechanism to transfer human-based BCI data to the training data. Table I shows the algorithm of the data preprocessing mechanism.

TABLE I. DATA PREPROCESSING MECHANISM ALGORITHM.

| |
|---|
| **Input:** |
| 1. $a_1, a_2, …, a_N, b_1, b_2, …, b_N, c_1, c_2, …, c_N, d_1, d_2, …, d_N, e_1, e_2, …, e_N$, and $t_1, t_2, …, t_N$<br>/*Parameters $a$, $b$, $c$, $d$, and $e$ are all of the collected human attention level, left-brain activation level, right-brain activation level, stress level, and fatigue whose values are in the interval [0, 10]. Parameter $t$ is the elapsed time after executing the developed psychological signal app and its unit is msec.*/ |
| 2. Each-move playing date and time for the game.<br>$datetime_{p1}, datetime_{p2}, …, datetime_{pM}$<br>/*Parameters $t_{p1}, t_{p2}, …, t_{pM}$ denote the playing date and time of move 1, 2, …, and $M$, respectively. |
| 3. $datetime_{ps}$ /*Starting date and time of executing the psychological signal app.*/ |

**Output:**
1. $a_{A1}, a_{A2}, ..., a_{AM}, b_{LBA1}, b_{LBA2}, ..., b_{LBAM}, c_{RBA1}, c_{RBA2}, ..., c_{RBAM}, d_{S1}, d_{S2}, ..., d_{SM}, e_{F1}, e_{F2}, ..., e_{FM}$
   /*The mapped data of the human attention level, left-brain activation level, right-brain activation level, stress level, and fatigue level for each move.*/
2. $a_{ALD1}, a_{ALD2}, ..., a_{ALDM}, b_{LBALD1}, b_{LBALD2}, ..., b_{LBALDM}, c_{RBALD1}, c_{RBALD2}, ..., c_{RBALDM}, d_{SLD1}, d_{SLD2}, ..., d_{SLDM}, e_{FLD1}, e_{FLD2}, ..., e_{FLDM}$
   /*The distance of the human's attention level, left-brain activation level, right-brain activation level, stress level, and fatigue level, for consecutive two moves.*/

**Method:**
**Step1:** For all elapsed time $t_i$, where $1 \leq i \leq N$
  **Step1.1:** Calculate the date and time of each record of brain signal
    $datetime_{bsi} = datetime_{ps} + t_{pi}$
**Step2:** For all $datetime_{bsi}$, where $1 \leq i \leq N$
  **Step2.1:** For all $datetime_{pj}$, where $1 \leq j \leq M$
    **Step 2.1.1:** If $datetime_{bsi} >= datetime_{pj}$
    Calculate the average of $a_i, b_i, c_i, d_i,$ and $e_i$ to acquire $a_{Aj}, b_{LBAj}, c_{RBAj}, d_{Sj}$, and $e_{Fj}$.
**Step3:** For all $a_{Ai}, b_{LBAi}, c_{RBAi}, d_{Si}$, and $e_{Fi}$, where $1 \leq i \leq M$
  **Step3.1:** Calculate the distance of human's attention level, left-brain activation level, right-brain activation level, stress level, and fatigue level for consecutive two moves.
**Step 4:** End

### C. Learning Process

This section describes the learning process, including an FML-based inference mechanism and a PFML-based learning mechanism. Table II shows the parameters of fuzzy sets and Table III shows partial knowledge base and rule base of the adopted FML, where there are six input fuzzy variables, including *Attention Level Distance* (*ALD*), *Brain Activation Level Distance* (*BALD*), *Stress Level Distance* (*SLD*), *Fatigue Level Distance* (*FLD*), *Simulations Number* (*SN*), and *Top-Move Rate* (*TMR*) as well as one output fuzzy variable *Win Rate* (*WR*).

TABLE II. PARAMETERS OF FUZZY SETS.

| *Attention Level Distance (ALD)* | | *Stress Level Distance (SLD)* | |
|---|---|---|---|
| Low | [0, 0, 0.5, 1] | Low | [0, 0, 0.5, 1] |
| Medium | [0.5, 1, 3, 4] | Medium | [0.5, 1, 3, 4] |
| High | [3, 4, 10, 10] | High | [3, 4, 10, 10] |
| *Brain Activation Level Distance (BALD)* | | *Fatigue Level Distance (FLD)* | |
| Low | [0, 0, 0.5, 1] | Low | [0, 0, 0.5, 1] |
| Medium | [0.5, 1, 3, 4] | Medium | [0.5, 1, 3, 4] |
| High | [3, 4, 10, 10] | High | [3, 4, 10, 10] |
| *Simulations Number (SN)* | | *Top-Move Rate (TMR)* | |
| Low | [0, 0, 128, 512] | Low | [0, 0, 0.8, 0.9] |
| High | [128, 512, 2048, 2048] | High | [0.8, 0.9, 1, 1] |
| *Win Rate (WR)* | | | |
| VeryLow | [0, 0, 0.35, 0.4] | High | [0.5, 0.6, 0.7, 0.8] |
| Low | [0.35, 0.4, 0.5, 0.6] | VeryHigh | [0.7, 0.8, 1, 1] |

TABLE III. PARTIAL KB AND RB OF THE ADOPTED FML.

```xml
<?xml version="1.0"?>
   <FuzzyController ip="localhost" name="">
     <KnowledgeBase>
       <FuzzyVariable domainleft="0" domainright="10" name="ALD" scale="" type="input">
         <FuzzyTerm name="Low" hedge="Normal">
           <TrapezoidShape Param1="0" Param2="0" Param3="0.5" Param4="1" />
         </FuzzyTerm>
         <FuzzyTerm name="Medium" hedge="Normal">
           <TrapezoidShape Param1="0.5" Param2="1" Param3="3" Param4="4" />
         </FuzzyTerm>
         <FuzzyTerm name="High" hedge="Normal">
           <TrapezoidShape Param1="3" Param2="4" Param3="10" Param4="10" />
         </FuzzyTerm>
       </FuzzyVariable>
                    :
     </KnowledgeBase>
   <RuleBase activationMethod="MIN" andMethod="MIN" orMethod="MAX" name="RuleBase1" type="mamdani">
     <Rule name="Rule1" connector="and" weight="1" operator="MIN">
       <Antecedent>
         <Clause>
           <Variable>ALD</Variable>
           <Term>Low</Term>
         </Clause>
         <Clause>
           <Variable>BALD</Variable>
           <Term>Low</Term>
         </Clause>
         <Clause>
           <Variable>SLD</Variable>
           <Term>Low</Term>
         </Clause>
         <Clause>
           <Variable>FLD</Variable>
           <Term>Low</Term>
         </Clause>
         <Clause>
           <Variable>SN</Variable>
           <Term>Low</Term>
         </Clause>
         <Clause>
           <Variable>TMR</Variable>
           <Term>Low</Term>
         </Clause>
       </Antecedent>
       <Consequent>
         <Clause>
           <Variable>WR</Variable>
           <Term>Low</Term>
         </Clause>
       </Consequent>
     </Rule>
                    :
     <Rule name="Rule324" connector="and" weight="1" operator="MIN">
       <Antecedent>
         <Clause>
           <Variable>ALD</Variable>
           <Term>High</Term>
         </Clause>
         <Clause>
           <Variable>BALD</Variable>
           <Term>High</Term>
         </Clause>
         <Clause>
           <Variable>SLD</Variable>
           <Term>High</Term>
         </Clause>
         <Clause>
           <Variable>FLD</Variable>
           <Term>High</Term>
         </Clause>
         <Clause>
           <Variable>SN</Variable>
           <Term>High</Term>
         </Clause>
         <Clause>
           <Variable>TMR</Variable>
           <Term>High</Term>
         </Clause>
       </Antecedent>
       <Consequent>
         <Clause>
           <Variable>WR</Variable>
           <Term>VeryHigh</Term>
         </Clause>
       </Consequent>
     </Rule>
   </RuleBase>
</FuzzyController>
```

We briefly describe them as follows: 1) *ALD*, *SLD*, and *FLD* are the attention level, stress level, and fatigue level distance of consecutive two moves, respectively. 2) *BALD* represents the brain activation level distance. Left-brain activation level and right-brain activation level are two corresponding indicators so here we use the distance of left-brain activation level as brain activation level distance. 3) *SN* is the FAIR ELF OpenGo-predicted number of simulations. 4) *TMR* denotes the matching degree of top-move rate prediction from ELF OpenGo AI bot [12]. 5) *WN* denotes the ELF OpenGo-predicted win rate of each move. The PFML-based learning mechanism combines particle swarm optimization and fuzzy markup language to learn the parameters of the fuzzy sets [8]. In this paper, there are 20 particles and the parameters of six input fuzzy variables and one output fuzzy variable represent the position of the particle in the seven dimensional space; they are optimized by adjusting the moving velocity in order to reach convergence. Additionally, the *inertia weight*, *cognitive parameter*, and *social parameter* of PSO are 0, 2, and 2, respectively. The fitness function *Fitness*($x_i, y_i$) is calculated as follows:

$$Fitness(x_i, y_i) = \sum_{i=1}^{M}(x_i - y_i)^2 / M \quad (1)$$

where *M* denotes the total number of the data points and $x_i$, and $y_i$ denote the inferred result and desired output of the *i*th data point, respectively. After termination of the learning, we use the positions of the best position among all 20 particles in the swarm to compose the after-learning knowledge base of the semantic BCI agent.

## IV. EXPERIMENTAL RESULTS

To evaluate the performance of the proposed approach, we invited four human Go players, including two professional players and two amateur players, to wear a mindo to play Go games. Table IV shows the basic information of the 10 games and the brief descriptions are in the following.

TABLE IV. INFORMATION OF 10 GAMES.

| Game No | Black | White | Winner |
|---|---|---|---|
| 1 | Hirofumi Ohashi (6P) | ELF OpenGo | W |
| 2 | Yi-Hsiu Lee (8P) | Hirofumi Ohashi (6P) | B |
| 3 | Darkforest | Yu-Hao Huang (2D) | B |
| 4 | Yi-Hsiu Lee (8P) + Robot (ELF OpenGo) | ELF OpenGo | B |
| 5 | Yi-Hsiu Lee (8P) | ELF OpenGo | W |
| 6 | Yi-Hsiu Lee (8P) | Hirofumi Ohashi (6P) | B |
| 7 | Yu-Hao Huang (2D) + Darkforest | Darkforest | W |
| 8 | Lu-An Lin (7D) + Robot (Darkforest) | Darkforest | W |
| 9 | ELF OpenGo | Lu-An Lin (7D) + Robot (ELF OpenGo) | B |
| 10 | ELF OpenGo | Lu-An Lin (7D) + Robot (ELF OpenGo) | B |

1) Games 1 and 5: Ohashi (6P) and Lee (8P) played with ELF OpenGo without any provided prediction information, respectively.
2) Games 2 and 6: Moves 1 to 88 of Games 2 and 6 were played by Ohashi (6P) and Lee (8P), respectively, according to the Moves 1 to 88 of Game 4. After the Move 88, they designed their own strategies and played the remaining moves in the Game 2 and Game 6 till the end of each game.
3) Game 3: Huang (2D) played against Darkforest without any provided prediction information.
4) Game 4: Lee (8P) acquired the robot's predicted next move advantage by listening and we set ELF OpenGo's simulations to 1024.
5) Game 7: Huang (2D) copied the Moves 1 to 88 of the Game 4 when playing the Game 7. After the Move 88, Huang (2D) played but also referred to the robot Palro's predicted moves until the end of the game.
6) Game 8: Lin (7D) and the robot Palro (Darkforest) were the team members of Black and they played against Darkforest by Pair Go.
7) Games 9 and 10: Lin (7D) and the robot Palro (ELF OpenGo) are the team members of Black and they played against ELF OpenGo as Pair Go. However, the difference between Games 9 and 10 is that Lin (7D) was allowed to refer to the robot's predicted moves before she played her next move.

Table V shows the game records and the human comments made by the two Go players. Figs. 5(a) and 5(b) show Lee (8P)'s psychological indicators of the Game 4 when he played the Move 67 and the whole Game 4, respectively. Figs. 6(a) and 6(b) are for the related information of the Game 4 played by Lee (8P) and Game 8 played by Lin (7D), respectively. Both had paid tremendous attention to the games. Lin (7D)'s stress level was also high stress in the whole game because her partner Darkforest put her at a disadvantage and she tried her best to turn the tables. For Game 8, Lee (8P)'s stress level changes a lot before about the Move 80. However, he experienced the high levels of stress for the last few moves before he won the game. Figs. 7(a)–(g) are the learned fuzzy sets for fuzzy variables *ALD*, *BALD*, *SLD*, *FLD*, *SN*, *TMR*, and *WR*, respectively, for learning 3000 generations. Fig. 8 shows the semantic accuracy before learning and after learning. It indicates that the proposed semantic BCI agent performs well after learning 3000 generations for most of the ten games except Game 2.

TABLE V. GAME RECORDS AND COMMENTS ON GAMES (A) 4 AND (B) 8.

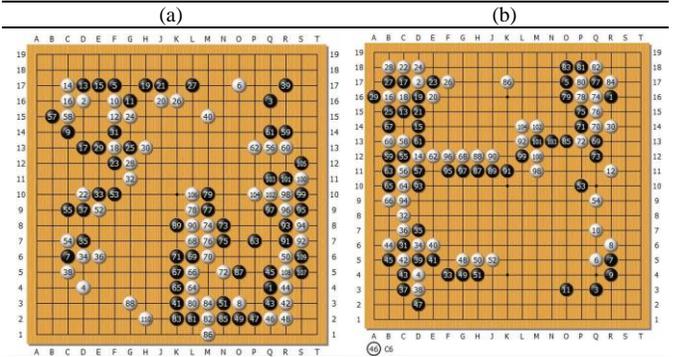

**Comments by Lee (8P)**
I took the Palro's suggestions to play for the first few moves so the situation was a draw. W64 is an over-played move so I launched a counter-attack but failed. Additionally, I played B73 at N8 instead of O9 because I misheard from the Palro. The win rate of Black started decreasing since B73. Finally, Black won the game because of White made a mistake from the Moves 91 to 94.

**Comments by Lin (7D)**
I once played against DF three years ago. At first, to defeat DF was a dream, but now beating DF was a breeze, after few rounds of practice. I was in partnership with DF as Black. Sometimes I ignored its suggested moves but sometimes I had to due to our partnership. Once my partner put me at a disadvantage and I tried my best to turn the tables. Since the Move 73, the game situation was not favorable for Black. Indeed, it was another kind of learning experience.

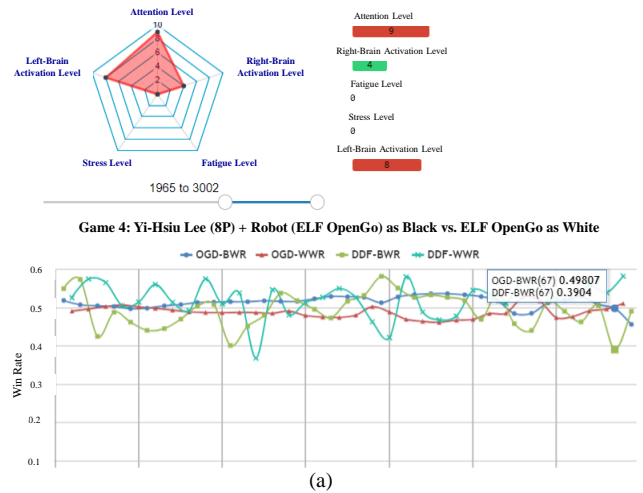

(a)

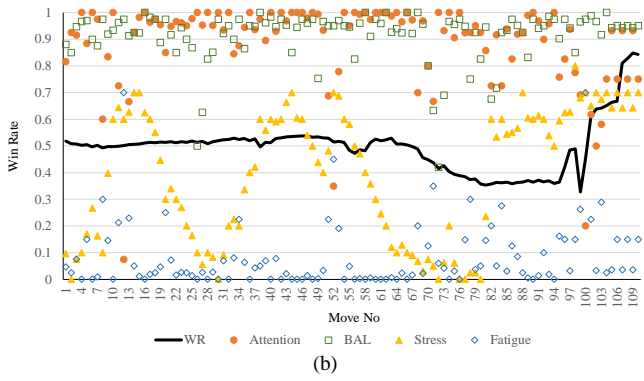

(b)

Fig. 5. Game 4: Lee (8P)'s psychological indicators and win rate of (a) move 67 and (b) all of the game.

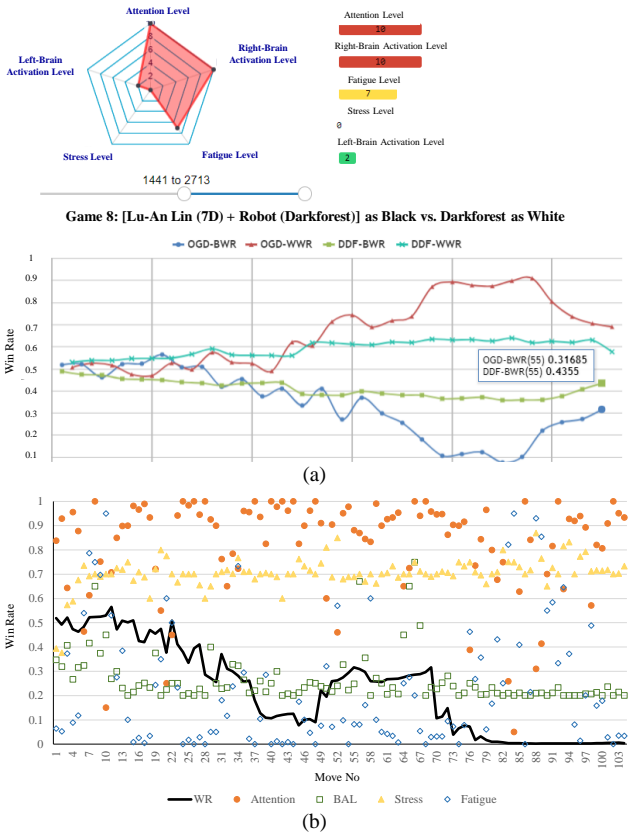

(b)

Fig. 6. Game 8: Lin (7D)'s psychological indicators and win rate of (a) move 55 and (b) all of the game.

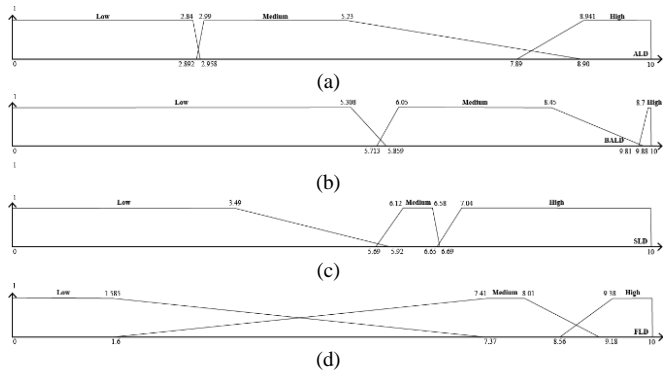

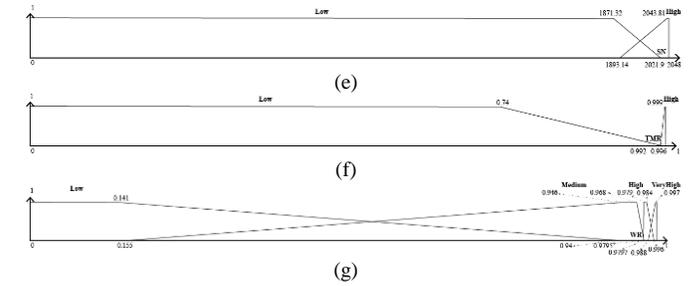

Fig. 7. After-learning fuzzy sets for fuzzy variables (a) *ALD*, (b) *BALD*, (c) *SLD*, (d) *FLD*, (e) *SN*, (f) *TMR*, and (g) *WR*.

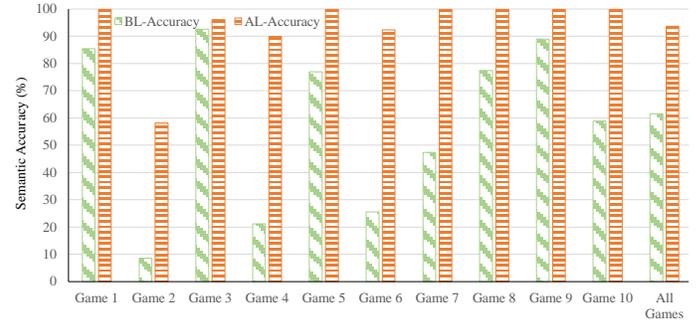

Fig. 8. Semantic accuracy.

## V. CONCLUSIONS

This paper proposes a semantic BCI agent based on FML and PSO for Go learning and prediction applications. We invited two professional Go players and two amateur Go players to join the experiment to retrieve the human-based BCI data. In each game, the machine-base game data are also stored in the repository. Based on these two kinds of data, we construct knowledge base and rule base to optimize the learned model based on FML and PSO. The experimental results show that the proposed human and smart machine co-learning mechanism create wonders. However, some weaknesses exist in the performance of the proposed method; therefore, improvements could be made, for example, by collecting more human-based BCI data to train much perfect model and by introducing the explainable intelligence into the developed OGD cloud platform.

### ACKNOWLEDGMENT

The authors would like to thank all the Go players who got involved in this research. We also would like to thank Wei-Chiao Chang, Yi-Chen Lu, all of the staff of KWS center & OASE Lab., and the Go club members at National Tainan First High School, including Yu-Han Huang and Yu-Lin Lin, to help collect the human-based BCI data. Additionally, we would like to thank Dr. Yuandong Tian and Facebook AI Research (FAIR) ELF OpenGo/Darkforest team members for their open source and technical support.

**Under review as a conference paper at IEEE CEC 2019**computer interfaces," *IEEE Systems, Man, & cybernetics Magazine*, vol. 3, no. 1, pp. 16–26, Oct. 2017.

[4] C. T. Lin, C. H. Chuang, C. S. Huang, S. F. Tsai, S. W. Lu, Y. H. Chen, and L. W. Ko, "Wireless and wearable EEG system for evaluating driver vigilance," *IEEE Transactoins on Biomedical Circuits and Systems*, vol. 8, no. 2, pp. 165–175, 2014.

[5] V. Martinez-Cagigal, E. Santamaria-Vazquez, J. Gomez-Pilar, and R. Hornero, "Towards an accessible use of smartphone-based social networks through brain-computer interfaces," *Expert Systes With Applications*, vol. 120, pp. 155–166, 2019.

[6] MarketsandMarkets Research Private Ltd, "AI in Education Market by Technology, Application, Component, Deployment, End-User, and Region - Global Forecast to 2023," Dec. 2018, [Online] https://www.marketsandmarkets.com/Market-Reports/ai-in-education-market-200371366.html.

[7] V. Matthews, "Teaching AI in schools could equip students for the future," May 2018, [Online] https://www.raconteur.net/technology/ai-in-schools-students-future.

[8] C. S. Lee, M. H. Wang, C. S. Wang, O. Teytaud, J. L. Liu, S. W. Lin, and P. H. Hung, "PSO-based fuzzy markup language for student learning performance evaluation and educational application," *IEEE Transactions on Fuzzy Systems*, vol. 26, no. 5, pp. 2618–2633, Oct. 2018.

[9] A. Edwards, C. Edwards, P. R. Spence, C. Harris, and A. Gambino, "Robots in the classroom: differences in students' perceptions of credibility and learning between teacher as robot and robot as teachers," *Compuers in Human Behavior*, vol. 65, pp. 627–634, 2016.

[10] C. S. Lee, M. H. Wang, T. X. Huang, L. C. Chen, Y. C. Huang, S. C. Yang, C. H. Tseng, P. H. Hung, and N. Kubota, "Ontology-based fuzzy markup language agent for student and robot co-learning," *2018 World Congress on Computational Intelligence* (*IEEE WCCI 2018*), Rio de Janeiro, Brazil, Jul. 8–13, 2018.

[11] Y. Tian and Y. Zhu, "Better computer Go player with neural network and long-term prediction," *2016 International Conference on Learning Representations* (*ICLR 2016*), San Juan, Puerto Rico, May 2–4, 2016 (https://arxiv.org/pdf/1511.06410.pdf).

[12] C. S. Lee, M. H. Wang, S. C. Yang, P. H. Hung, S. W. Lin, N. Shuo, N. Kubota, C. H. Chou, P. C. Chou, and C. H. Kao, "FML-based dynamic assessment agent for human-machine cooperative system on game of Go," *International Journal of Uncertainty, Fuzziness and Knowledge-Based Systems*, vol. 25, no. 5, pp. 677–705, 2017.

[13] Y. Tian and L. Zitnick, "Facebook Open Sources ELF OpengGo," May 2018, [Online] Available: https://research.fb.com/facebook-open-sources-elf-opengo/.

[14] C. S. Lee, M. H. Wang, L. W. Ko, N. Kubota, L. A. Lin, S. Kitaoka, Y. T Wang, and S. F. Su, "Human and smart machine co-learning: brain-computer interaction at the 2017 IEEE International Conference on Systems, Man, and Cybernetics," *IEEE Systems, Man, and Cybernetics Magazine*, vol. 4, no. 2, pp. 6–13, Apr. 2018.

[15] D. Silver, J. Schrittwieser, K. Simonyan, I. Antonoglou, A. Huang, A. Guez, T. Hubert, L. Baker, M. Lai, A. Bolton, Y. Chen, T. Lillicrap, F. Hui, L. Sifre, G. v. d. Driessche, T. Graepel, and D. Hassabis, "Mastering the game of Go without human knowledge," *Nature*, vol. 550, pp. 35–359, 2017.

[16] R. Sitaram, S. Lee, S. Ruiz, M. Rana, R. Veit, and N. Birbaumer, "Real-time support vector classification and feedback of multiple emotional brain states," *Neuroimage.*, vol. 56, pp. 753–765, May 2011.

[17] K. P. Thomas and A. P. Vinod, "EEG-based biometric systems," *IEEE Systems, Man, & cybernetics Magazine*, vol. 3, no. 1, pp. 6–15, Oct. 2017.

[18] G. Acampora, V. Loia, C. S. Lee, and M. H. Wang, "On the Power of Fuzzy Markup Language," Springer-Verlag, Germany, Jan. 2013.

[19] IEEE Standards Association, "1855-2016 - IEEE Standard for Fuzzy Markup Language," May 2016, [Online] Available: http://ieeexplore.ieee.org/document/7479441/?arnumber=7479441&filter=AND(p_Publication_Number:7479439).

[20] Z. Zeng, C. Miao, C. Leung, and C. J. Jih, "Building more explainable artificial intelligence with argumentation," *Thirty-Second AAAI Conference on Artificial Intelligence* (*AAAI 2018*), New Orleans, USA, Feb. 2–7, 2018.

[21] C. S. Lee, M. H. Wang, M. J. Wu, O. Teytaud, and S. J. Yen, "T2 FS-based adaptive linguistic assessment system for semantic analysis and human performance evaluation on game of Go," *IEEE Transactions on Fuzzy Systems*, vol. 23, no. 2, pp. 400–420, Apr. 2015.

[22] C. S. Lee, M. H. Wang, and S. T. Lan, "Adaptive personalized diet linguistic recommendation mechanism based on type-2 fuzzy sets and genetic fuzzy markup language," *IEEE Transactions on Fuzzy Systems*, vol. 23, no. 5, pp. 1777–1802, Oct. 2015.

[23] L. Sanders, ScienceNews, "Brain waves may focus attention and keep information flowing," Mar. 2018, [Online] Available: https://www.sciencenews.org/article/brain-waves-may-focus-attention-and-keep-information-flowing.

[24] Brain Rhythm INC., Dec. 2018, [Online] Available: http://www.bri.com.tw/product_br8plus.html.

[25] Anatomy of the brain, Mayfield Brain & Spine, Apr. 2018, [Online] Available: http://www.mayfieldclinic.com/PE-AnatBrain.htm.

[26] L. W. Ko, O. Komarov, W. D. Hairston, T. P. Jung, and C. T. Lin, "Sustained attention in real classroom settings: An eeg study," *Frontiers in human neuroscience*, vol. 11, no. 388, 2017.